\documentclass[letterpaper, 10 pt, conference]{ieeeconf}
\IEEEoverridecommandlockouts 
\usepackage{amsmath,amssymb,amsfonts}
\usepackage{xcolor}
\usepackage{graphicx}
\usepackage{xspace}
\usepackage[ruled,vlined,noend]{algorithm2e}
\usepackage{siunitx}
\usepackage{multirow}
\usepackage[normalem]{ulem}
\usepackage{hyperref}

\SetCommentSty{mycommfont}
\newcommand{\inputstate}{\mathbf{s}}
\newcommand{\network}{\mathcal{N}}
\newcommand{\PILOT}{\textsc{PILOT}\xspace}
\newcommand{\obp}{\mbox{\textsc{2s-OPT}}\xspace}
\newcommand{\networkobp}{\network_\theta^{\text{2s-OPT}}}

\newcommand{\x}{\mathbf{x}}
\newcommand{\cnt}{\mathbf{u}}

\title{\LARGE \bf PILOT: Efficient Planning by Imitation Learning and Optimisation\\for Safe Autonomous Driving}

\author{\authorblockN{Henry Pulver\authorrefmark{1}, Francisco Eiras\authorrefmark{1}\authorrefmark{2}, Ludovico Carozza\authorrefmark{1}, Majd Hawasly\authorrefmark{1}, \\Stefano~V.~Albrecht\authorrefmark{1}\authorrefmark{3} and Subramanian Ramamoorthy\authorrefmark{1}\authorrefmark{3}}
\authorblockA{\authorrefmark{1}Five\,AI Ltd., United Kingdom
\\Email: \texttt{first.last@five.ai}}
\authorblockA{\authorrefmark{2}University of Oxford, United Kingdom} \authorblockA{\authorrefmark{3}University of Edinburgh, United Kingdom}
}

\begin{document}

\maketitle

\begin{abstract}
Achieving a proper balance between planning quality, safety and efficiency is a major challenge for autonomous driving. 
Optimisation-based motion planners are capable of producing safe, smooth and comfortable plans, but often at the cost of runtime efficiency. On the other hand, na\"ively deploying trajectories produced by efficient-to-run deep imitation learning approaches might risk compromising safety.
In this paper, we present \PILOT -- a planning framework that comprises an imitation neural network followed by an efficient optimiser that actively rectifies the network's plan, guaranteeing fulfilment of safety and comfort requirements. The objective of the efficient optimiser is the same as the objective of an expensive-to-run optimisation-based planning system that the neural network is trained offline to imitate. This efficient optimiser provides a key layer of online protection from learning failures or deficiency in out-of-distribution situations that might compromise safety or comfort. 
Using a state-of-the-art, runtime-intensive optimisation-based method as the expert, we demonstrate in simulated autonomous driving experiments in CARLA that \PILOT achieves a seven-fold reduction in runtime when compared to the expert it imitates without sacrificing planning quality.

\end{abstract}

\section{Introduction}
Guaranteeing safety of decision-making is a fundamental challenge on the path towards the long-anticipated adoption of autonomous vehicle (AV) technology. Attempts to address this challenge show the diversity of possible approaches to the concept of safety: whether it is maintaining the autonomous system inside a safe subset of possible future states~\cite{batkovic2019real, chen2019autonomous}, preventing the system from breaking domain-specific constraints~\cite{schwarting2017safe, obp}, or exhibiting a behaviour that matches the safe behaviour of an expert~\cite{sadat2020perceive}, amongst others.

\begin{figure}[t]
    \centering
    \includegraphics[width=0.48\textwidth]{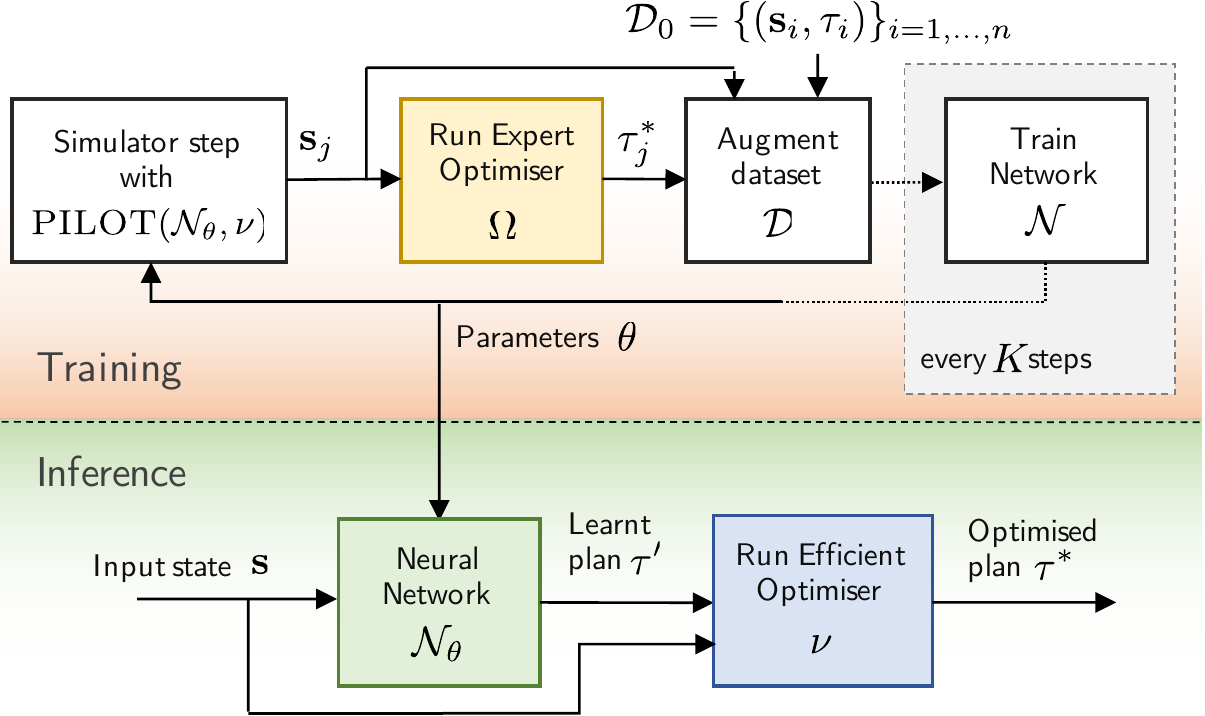}
    \caption{\textit{\PILOT framework}: (top) \PILOT uses an expert-in-the-loop imitation learning paradigm to train a deep neural network, $\network_\theta$, that imitates the output of an expensive-to-run optimisation-based planner $\Omega$. (bottom) At inference time, \PILOT  uses the output of $\network_\theta$ to initialise an efficient optimiser $\nu$ to compute a feasible and low cost trajectory.}
    \label{fig:general-pilot}
    \vspace{-0.8em}
\end{figure}

\begin{figure}[t]
    \centering
    \vspace{0.25cm}
    \includegraphics[width=0.48\textwidth]{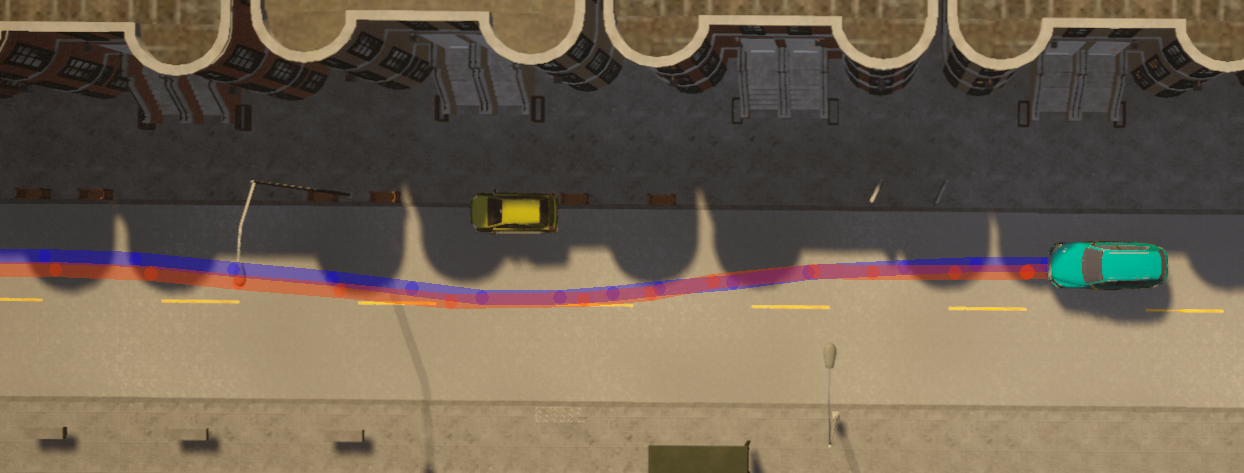}
    \caption{An example CARLA scenario with trajectories generated with the planner-in-the-loop with a horizon of 8s per planning stage for 1) the expensive-to-run planner \obp (in red) that took $\SI{175}{\milli\second}$ per planning stage on average, and  2) \PILOT (in blue) that took $\SI{44}{\milli\second}$ per planning stage on average. More examples in the accompanying video \tt{\href{https://www.five.ai/pilot}{five.ai/pilot}}}
    \label{fig:example}
    \vspace{-1.5em}
\end{figure}

Approaches to motion planning in AVs can be categorised in different ways, e.g., \textit{data-driven} vs. \textit{model-based}. 
The hands-off aspect of purely \textit{data-driven} approaches is lucrative, which is evidenced by the growing interest in the research community in exploiting techniques such as reinforcement or imitation learning applied to autonomous driving~\cite{bojarski2016end, pan2020imitation, hawke2020urban, chen2020learning}. 
Moreover, inference in a data-driven model is usually efficient when compared to more elaborate search- or optimisation-based approaches, which is a key requirement in real-time applications. However, this does not come for free as these systems struggle to justify their decision making or to certify the safety of their output at deployment time without major investments in robust training~\cite{mirman2018differentiable, parot} or post-hoc analysis~\cite{liu2019algorithms}. This constitutes a major setback to the deployment of such methods in safety-critical contexts.
On the other hand, \textit{model-based} approaches tend to be engineering-heavy and require deep knowledge of the application domain, while giving a better handle on setting and understanding system expectations through model specification. Moreover, they produce more interpretable plans~\cite{hanna21, albrecht2020integrating,DeCastro2020InterpretablePF,paden2016survey}. This, however, usually comes at the cost of robustness~\cite{schwarting2017safe} or runtime efficiency~\cite{obp}. We aim to bring the efficiency benefits of data-driven methods together with the guarantees of model-based systems in a \textit{hybrid} approach for urban driving applications. Our goal is to introduce a general planning architecture that is flexible enough to capture complex planning requirements, yet still guarantee the satisfaction of sophisticated specifications at deployment, without sacrificing runtime efficiency.

In this work, we propose an approach that combines model-based optimisation and deep imitation learning, using as the expert a performant optimisation-based planner that is expensive to run. We introduce \textbf{\PILOT} -- \textit{Planning by Imitation Learning and Optimisation} -- in Sec.~\ref{sec:pilot}.  
At \textit{training time} (Fig.~\ref{fig:general-pilot}, top), we \textit{distil}~\cite{policy-distillation} the expert planner's behaviour using imitation learning, with \textit{online}, expert-in-the-loop  dataset augmentation  (e.g. DAgger -- Dataset Aggregation~\cite{dagger}) to continually enrich the training dataset with relevant problems sampled from the state distribution induced by the learner's policy. At \textit{inference time} (Fig.~\ref{fig:general-pilot}, bottom), to actively correct potential learning failures and improve safety, we employ an efficient optimisation component that optimises the same objective function as the expert but benefits from informed warm-starting provided by the network output. 

In this paper, without loss of generality to our approach,  we use the two-stage optimisation framework introduced by Eiras \emph{et al.}~\cite{obp} as the expensive-to-run planner to imitate in a simulated environment. As discussed by the authors, the framework in~\cite{obp}   
suffers in terms of runtime, effectively trading off efficiency for better solution quality, which makes it a suitable choice for \PILOT. We use the CARLA simulator~\cite{carla} to validate our approach under a wide variety of conditions in realistic simulations. This is an important step in the direction of understanding  the viability and safety of such methods before deploying on the roads. A qualitative example of \PILOT's performance in CARLA is in Fig.~\ref{fig:example}.

The contributions of this work are:
\begin{itemize}
    \item A robust and scalable framework that imitates an expensive-to-run optimiser, with expert-in-the-loop data augmentation at training time and active correction at inference time using an efficient optimiser.
    \item Applying this framework to the two-stage optimisation-based planner from \cite{obp} leading to a $7\times$ runtime improvement in our benchmark CARLA  datasets at no significant loss in solution quality (measured by the  objective function cost of the output trajectory). 
\end{itemize}

\section{Background and Related Work}
\label{sec:background}

In this section we review related work in motion planning for AVs regarding imitation learning with optimisation experts  (Sec.~\ref{sec:il-ad}) and motion planning via optimisation  (Sec.~\ref{sec:planning-in-opt-background}). Then, we give an overview of a planning method we use in this work to demonstrate \PILOT for (Sec.~\ref{sec:obp-details}).

\subsection{Imitation Learning with Optimisation Experts for AVs}
\label{sec:il-ad}
With the complexity of specifying the objective function of safe, assertive driving, imitation learning offers a promising alternative. However, na\"ive attempts to leverage expert traces, e.g. with vanilla behavioural cloning~\cite{pomerleau1989alvinn}, usually fails at deployment to exhibit safe behaviour   in complex scenarios due to covariate shift between the training and deployment settings~\cite{codevilla2019exploring, filos2020can}. To mitigate this issue,  techniques for training data augmentation range from online methods that actively enrich the training dataset with actual experiences from the deployment environment~\cite{dagger}, and offline synthesis of realistic scenarios for the expert to demonstrate recovery from perturbations~\cite{chen2019deep} or near-misses~\cite{bansal2018chauffeurnet}.

Still, data augmentation by itself cannot guarantee the safety of decisions at inference time, which we believe is a fundamental requirement for any deployed system in the safety-critical autonomous driving task. Yet, most of the existing literature on imitation learning of optimisation~\cite{pan2020imitation, lee2018safe, sun2018fast,acerbo2020safe} propose pure end-to-end learning pipelines. 

Pan \textit{et al.} in~\cite{pan2020imitation} proposed an end-to-end system for off-road, fixed route, real-world planning that learns to map basic sensory input into controls with the guidance of a Model Predictive Control (MPC) expert that has access to better sensors and more compute. However, no safety guarantees are provided at deployment time beyond what a low-level controller employed to track the network output does. 
A related approach by Sun \textit{et al.} in~\cite{sun2018fast} employs a shallow neural network with selected state features as input to imitate an MPC expert that optimises progress and control effort in long-term, two-lane driving scenarios with two other vehicles. On top of that, an online, short-horizon MPC controller tracks the initial portion of the inferred trajectory, constrained by the same feasibility and collision constraints. For online augmentation, the optimisation problems in which the network output deviates away from the expert's are included in the augmented training dataset. Acerbo \textit{et al.} in~\cite{acerbo2020safe} pre-train a fully-connected neural network to map state features of a simple lane-keeping scenario involving no other vehicles into parameters of smooth, second-order polynomial curves, using a dataset generated by a short-horizon nonlinear MPC expert. In addition to the usual $L_2$ term, the training loss incorporates other terms related to collision avoidance, implemented with barrier functions. A related,  supervised learning approach is  Constrained Policy Nets~\cite{cpn} in which the loss of a policy network is derived directly from an optimisation objective. This, however, requires careful treatment of the constraint set to ensure differentiability.

Another approach to rectify an imitation network output   employs control safe sets to validate acceleration and steering commands  of an imitation network  trajectory~\cite{chen2019deep}. This, however, is limited to taming the predicted trajectory inside the safe set, but unable to suggest other viable corrections.

\begin{figure*}[t]
    \centering
    \includegraphics[width=0.99\textwidth]{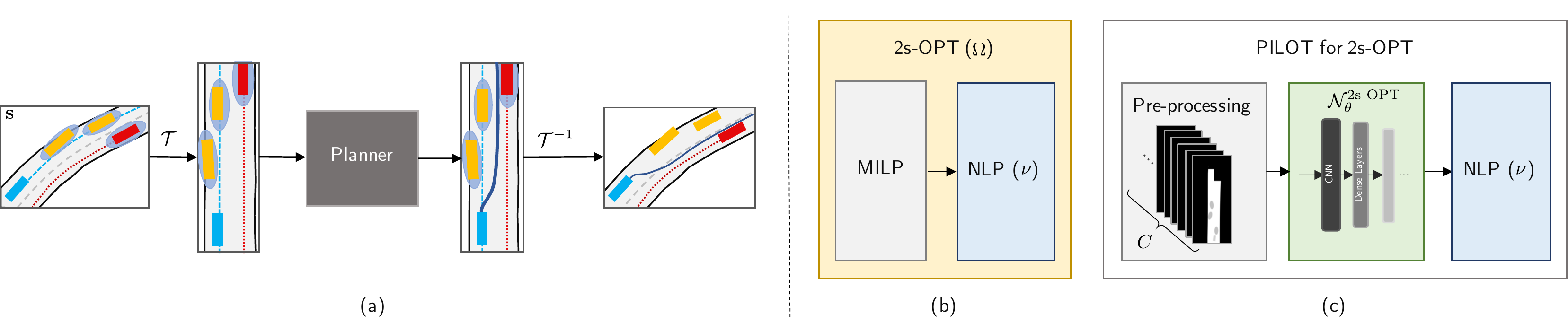}
    \vspace{-1.0em}
    \caption{\textit{\PILOT for \obp}: (a) The planning input state is first transformed 
    from the global coordinate frame to the reference path-based coordinate frame. After the plan is obtained, the output is transformed back to the global coordinate frame by the inverse transform.  
    See~\cite{obp} for more details; (b) Architecture of \obp (expert,  expensive-to-run planner $\Omega$): a MILP solver initialises an NLP optimiser; (c) Architecture of \PILOT  for \obp: input pre-processing produces a sequence of images of the scene, encoding road surface information and the predicted future of dynamic road users, the network $\networkobp$ which was specifically designed for this problem, and the NLP problem as in \obp ($\nu$).}
    \label{fig:architecture}
    \vspace{-0.8em}
\end{figure*}

\subsection{Motion Planning as Optimisation for AVs}
\label{sec:planning-in-opt-background}

For any variable $v_j$ with $j \in \mathbb{Z}^*$, we will use the shorthand $v_{i:e} = \{v_i,...,v_e\}$. 
In its most general form, motion planning via optimisation is defined as follows~\cite{paden2016survey}: assume the input to the motion planning problem is given by a scene $\inputstate\in \mathcal{S}$ (vehicle states/positional uncertainty, layout information, and predictions of other agents over a fixed horizon). The goal is to obtain a plan for the ego-vehicle (or \textit{ego} for short) as a sequence of $N+1$ states, $\tau^* = \tau_{0:N} \in \mathcal{T}$, such that:
\begingroup\makeatletter\def\f@size{10}\check@mathfonts
\begin{equation}
\begin{aligned}
& \tau^* =\underset{\tau}{\text{argmin}}\; \mathcal{J}(\tau) \\
& \text{\hspace{1.3em} s.t.}\; \tau_0 = \inputstate_{\text{ego}}\text{, }\;\; \mathbf{g}_\inputstate(\tau_{0:N}) \leq 0\text{, }\;\;\mathbf{h}_\inputstate(\tau_{0:N}) = 0\\
\end{aligned}
\label{eq:general_problem}
\end{equation}
\endgroup
where $\mathcal{J}$ is a cost function of progress and comfort terms defined over the plan $\tau$, $\inputstate_{\text{ego}}$ refers to the initial \textit{ego} state in the scene $\inputstate$, $\mathbf{g}_\inputstate$ and $\mathbf{h}_\inputstate$ are sets of general inequality and equality constraints, respectively, parameterised by the input scene on the ego-vehicle states. These constraints typically ensure the satisfaction of strict requirements of model dynamics and safety (up to the predicted horizon) of the output plans~\cite{paden2016survey,schwarting2017safe,obp}.

There is a wide-ranging literature on attempts to solve relaxations of the problem in Eq.~\ref{eq:general_problem}, e.g. by turning it into unconstrained optimisation, taking a convex approximation, tackling simplified driving scenes, or limiting the planning horizon~\cite{paden2016survey,sun2018fast,acerbo2020safe}. A recent work by Schwarting \textit{et al.} solves Eq.~\ref{eq:general_problem} directly in a receding horizon fashion~\cite{schwarting2017safe}, yet the authors identify local convergence as a setback of their method\footnote{While the  cost function in that work is for semi-autonomous driving, this does not affect the general difficulty of the problem.}. In~\cite{obp}, Eiras \textit{et al.}  mitigate this issue by warm-starting the solver, which negatively affects runtime efficiency. We describe next the architecture of~\cite{obp}, which we then use in Sec.~\ref{sec:pilot-for-obp} to practically demonstrate the effectiveness of \PILOT.

\subsection{A Two-stage Optimisation-based Motion Planner}
\label{sec:obp-details}

Fig~\ref{fig:architecture}(a, b) show the general architecture of the two-stage optimiser of~\cite{obp}, which we will refer to as \textbf{\obp}. The input to the system are: 1) a birds-eye view of the planning scene, that includes the {ego-vehicle}, other road users and the relevant features of the static layout; 2) a  reference route provided by an external route planner; and 3) predicted traces for all road users provided by a prediction module. Projecting the world state and predictions into a reference path-based coordinate frame produces the \obp input (Fig~\ref{fig:architecture}(a)).

The Nonlinear Programming (NLP) problem solved in~\cite{obp} follows the structure of Eq.~\ref{eq:general_problem} with the following constraints:    1) \textit{Kinematic feasibility} (equality): an \textit{ego} state at time $k$ can be obtained by applying a discrete bicycle model to the state at time $k-1$;
    2) \textit{Velocity limits} (inequality): the speed is lower-bounded by the minimum speed (typically 0) and upper-bounded by the speed limit; 
    3) \textit{Control input bounds} (inequality): the control inputs are lower- and upper-bounded; 
    4) \textit{Jerk bounds} (inequality): the change in the control inputs is lower- and upper-bounded; 
    5) \textit{Border limits} (inequality): the \textit{ego} remains within the driveable surface (e.g., the road surface, or lane if overtaking is not allowed);
    6) \textit{Collision avoidance} (inequality): the \textit{ego} does not collide with any other road user or object.

The cost function  $\mathcal{J}_{\text{2s-OPT}}$  comprises a linear combination of quadratic terms of comfort (bounded acceleration and jerk) and progress (longitudinal and lateral tracking of the reference path, as well as target speed). More details on the precise formulation of the constraints, cost function and parameters are available in Appendix~\ref{apx:nlp}.

To solve this optimisation problem, the two-stage architecture presented in Fig~\ref{fig:architecture}(b) is applied. The first stage solves a receding-horizon, linearised version of the planning problem using a Mixed-Integer Linear Programming (MILP) solver. The output of the MILP stage is fed in one go as a warm-start initialisation to the NLP optimiser. This second optimisation stage ensures that the output trajectory is smooth and feasible, while maintaining safety guarantees.

\section{\PILOT: Planning by Imitation Learning and Optimisation}
\label{sec:pilot}

We now introduce \PILOT, an efficient general solution to attain the benefits of expensive-to-run optimisation-based planners. 
As is well known in the community, while  solving the general problem in Eq.~\ref{eq:general_problem} globally is NP-hard~\cite{floudas2013state, nocedal2006numerical}, there are efficient solvers that can compute local solutions  within acceptable times \textit{if} a sensible initialisation is provided~\cite{ipopt,forcesnlp}. We define $\nu: \mathcal{S}\times \mathcal{T} \to \mathcal{T}$ to be such an \textit{efficient optimiser}. We denote by $\Omega: \mathcal{S}\to \mathcal{T}$ the `expert', \textit{expensive-to-run optimisation} procedure that has the potential to converge from an uninformed initialisation to lower cost solutions than $\nu$.
Practical examples of $\Omega$ include recursive decompositions of the problem and taking the minimum cost~\cite{friesen2016recursive}, or informed warm-starting~\cite{obp, lembono2020memory}.

\begin{algorithm}[t]
    \DontPrintSemicolon
    \SetAlgoLined
    \SetNoFillComment
    \LinesNotNumbered 
    \SetKwInOut{Input}{input}
    \SetKwInOut{Output}{output}
    
    \Input{state $\mathbf{s}$, trained imitation network $\mathcal{N}_\theta$, efficient planner $\nu$}
    \Output{optimal plan $\tau^*$}
    Obtain initial trajectory $\tau_{\mathcal{N}_\theta} \leftarrow \mathcal{N}_\theta(\mathbf{s})$\;
    Get $\tau^*$ by optimising $\mathcal{J}$ using $\nu(\mathbf{s},\tau_{\mathcal{N}_\theta})$\;
    \Return{$\tau^*$}
    \caption{\textsc{PILOT Inference Step}}
    \label{alg:inference}
\end{algorithm}

The goal of \PILOT is to safely achieve low costs on $\mathcal{J}$ comparable to the ones achievable by $\Omega$, in runtimes comparable to the efficient $\nu$. To do so, \PILOT employs an imitation learning paradigm to train a deep neural network $\network_\theta$ to imitate the output of $\Omega$, which it then uses at inference time to initialise $\nu$. 
While in theory the network would naturally achieve a low cost while satisfying the constraints (perfect learning), in practice this is not the case. As such, $\nu$ works as an efficient online correction mechanism that uses this informed initialisation to output low cost, safe and feasible trajectories. More details about the inference procedure is shown in Algorithm~\ref{alg:inference} and  Fig.~\ref{fig:general-pilot}~(bottom).
%

In order to achieve that, we pre-train the network on problems solved by the expert planner $\Omega$, $\mathcal{D}_0 = \{(\inputstate_i, \tau^*_i)\}_{i=1:n}$. Then, with the pre-trained network $\mathcal{N}_\theta$ and the efficient optimiser $\nu$ acting as a planner, we employ a DAgger-style training loop~\cite{dagger} in a simulator to adapt to the covariate shift in $\mathcal{D}_0$ to the learner's experience in the simulator. For more details about training, see Algorithm~\ref{alg:training}  and Fig.\ref{fig:general-pilot}~(top).

\begin{algorithm}[t]
    \DontPrintSemicolon
    \SetAlgoLined
    \SetNoFillComment
    \LinesNotNumbered 
    \SetKwInOut{Input}{input}
    \SetKwInOut{Output}{output}
    
    \Input{initial dataset $\mathcal{D}_0 = \{(\mathbf{s}_i, \tau^*_i)\}_{i=1:n}$, expert planner $\Omega$, efficient planner $\nu$, simulator $\mathcal{S}$, training problems count $J$, retrain count $K$}
    \Output{trained network parameters $\theta$}
    Initialise $\mathcal{D}$ to $\mathcal{D}_0$\;
    Pre-train $\theta \leftarrow \text{\textsc{Train}}(\network, \mathcal{D})$\;
    \For{$j \in \{n+1, \ldots, J\}$}{
    
        \eIf{simulation finished}{
            $\mathbf{s'} \leftarrow$ Initialise a new simulation}
        {$\mathbf{s'} \leftarrow \mathbf{s}_{j-1}$}
    
        Obtain $\mathbf{s}_j$ from $ \mathcal{S}$ by  $\text{\textsc{PILOT}}(\mathbf{s'}; \mathcal{N}_{\theta}, \nu)$ step\;
        Get $\tau_j^*$ by optimising $\mathcal{J}$ using $\Omega(\mathbf{s}_j)$\;
        Update $\mathcal{D} \leftarrow \mathcal{D} \cup \{(\mathbf{s}_j, \tau_j^*)\}$\;
        \tcp*[l]{retrain network every $K$ steps}
        \If{$(j-n)\bmod K = 0$}{
            Update $\theta \leftarrow \text{\textsc{Train}}(\mathcal{N}, \mathcal{D})$\;
        }
    \,
    }
    \Return{$\theta$}
    \caption{\textsc{PILOT Training Procedure}}
    \label{alg:training}
\end{algorithm}

\section{\PILOT for the Two-stage Optimisation-based Motion Planner}
\label{sec:pilot-for-obp}

To demonstrate the effectiveness of \PILOT, we apply it to the use case of \obp. To do so, we take \obp as the expensive-to-run planner $\Omega$, and borrow its NLP constrained optimisation stage as the efficient optimisation planner $\nu$ -- see Fig.~\ref{fig:architecture}(c).
We design a deep neural network $\networkobp$ that outputs smooth trajectories given as input a graphical representation of the scene and other scalar parameters of the problem (e.g. ego-vehicle speed). We train the network using Algorithm~\ref{alg:training} to imitate the output of \obp when presented with the same planning problem.

The planning scene, $\mathbf{s}$, comprises the static road layout, road users with predicted trajectories, and a reference path to follow which acts as a behaviour conditioning input (Fig.~\ref{fig:architecture}(a)). As the problem is transformed to the reference path coordinate frame, the resulting scene is automatically  aligned with the area of interest -- the road along the reference path, simplifying the network representation.

To graphically encode the predicted trajectories of dynamic road users, $C$ greyscale, top-down images of the scene $I_t^{\mathbf{s}} \in \mathbb{R}^{W\times H}$ are produced by sampling the predicted positions of road users uniformly at times $t~\in~\{0, \frac{h}{C-1}, \ldots, h\}$, for a planning horizon $h = N\Delta t$. These images are stacked $\mathcal{I}^{\mathbf{s}} = I_{1:C}^{\mathbf{s}} \in \mathbb{R}^{C\times W\times H}$ and fed into convolutional layers to extract semantic features, as shown in Fig.~\ref{fig:architecture}(c). This is similar to the input representation in previous works, e.g. ChauffeurNet~\cite{bansal2018chauffeurnet}, with the exception that in our case the static layout information is repeated on all channels.

Additional information of the planning problem that is not visualised in the top-down images (such as the initial speed of the ego-vehicle) is appended as scalar inputs  along with the flattened convolutional layers output  to the first dense layer of the network. Refer to Appendix~\ref{sec:app_network_architecture} for more details.

The desired output of the network is a trajectory in the reference path coordinate frame, encoded as a vector of time-stamped positions  $\rho^\theta~=~\{(x_j, y_j)\}_{j=1,...,N} \in \mathbb{R}^{2\times N}$. With this representation, we define the training loss to be the $L_2$ norm between the expert trajectory and the network output:
\begin{equation}
\label{eq:loss_fn}
\mathcal{L}_{\theta}(\mathcal{D}) = \frac{1}{nN}\sum_{i \in \mathcal{D}} ||\rho_i^\theta - \rho_i^*||^2 + \mu ||\theta||^2,
\end{equation}
where $\theta$ is the neural network parameter vector, $\mathcal{D}$ is the training dataset, $\rho_i^*$ is the expert's time-stamped position at index $i$ from the dataset, and $\mu$ is a regularisation parameter.

The efficient NLP optimisation planner (Sec.~\ref{sec:obp-details}) expects as initialisation a time-stamped  sequence of positions, speeds, orientations and control inputs (steering and acceleration) over the full-horizon, all in the reference path coordinate frame, as a single optimisation problem (cf. the traditional receding-horizon setting). We calculate speeds and orientations from the network's output sequence (after post-processing -- Appendix~\ref{apx:checks}), and derive the control values from the inverse dynamics model.

\section{Experiments}
\label{sec:experiments}

In this section, we attempt to answer the following questions to demonstrate the effectiveness of \PILOT: 
\begin{enumerate}
    \item \textit{How does \PILOT fare compared to the expert, expensive-to-run optimiser it is trained to imitate?}
    \item \textit{Is the imitation neural network alone  sufficient to produce safe, feasible and low cost solutions, similar to those of the expert?}
    \item \textit{Is the imitation network necessary for the efficient optimiser $\nu$ to converge to feasible and low cost solutions, or are simple heuristics sufficient?}
    \item \textit{How does \PILOT compare to a baseline that trains a network to directly optimise the objective of  $\nu$?}
\end{enumerate}

To answer question 1), we compare \PILOT and \obp in terms of solving time and closed-loop trajectory cost using $\mathcal{J}_{\text{2s-OPT}}$ (Sec.~\ref{sec:q1}). We investigate question 2) by comparing constraint satisfaction in \PILOT and in the imitation network $\networkobp$ alone (Sec.~\ref{sec:q2}). To shed light on question 3) we perform an ablation on the initialisation of the efficient optimiser, in this case the NLP solver, by swapping the network with different heuristics and comparing the solution quality (Sec.~\ref{sec:q3}). Finally to answer question 4), we implement the state-of-the-art Constrained Policy Net (CPN)~\cite{cpn}, in which a neural network is trained directly with a loss function that approximates the optimiser's objective, and compare it to \PILOT with regard to constraint satisfaction (Sec.~\ref{sec:q4}).

\subsection{Experimental Setup}

\begin{figure}[t]
\centering
 \includegraphics[width=0.45\textwidth]{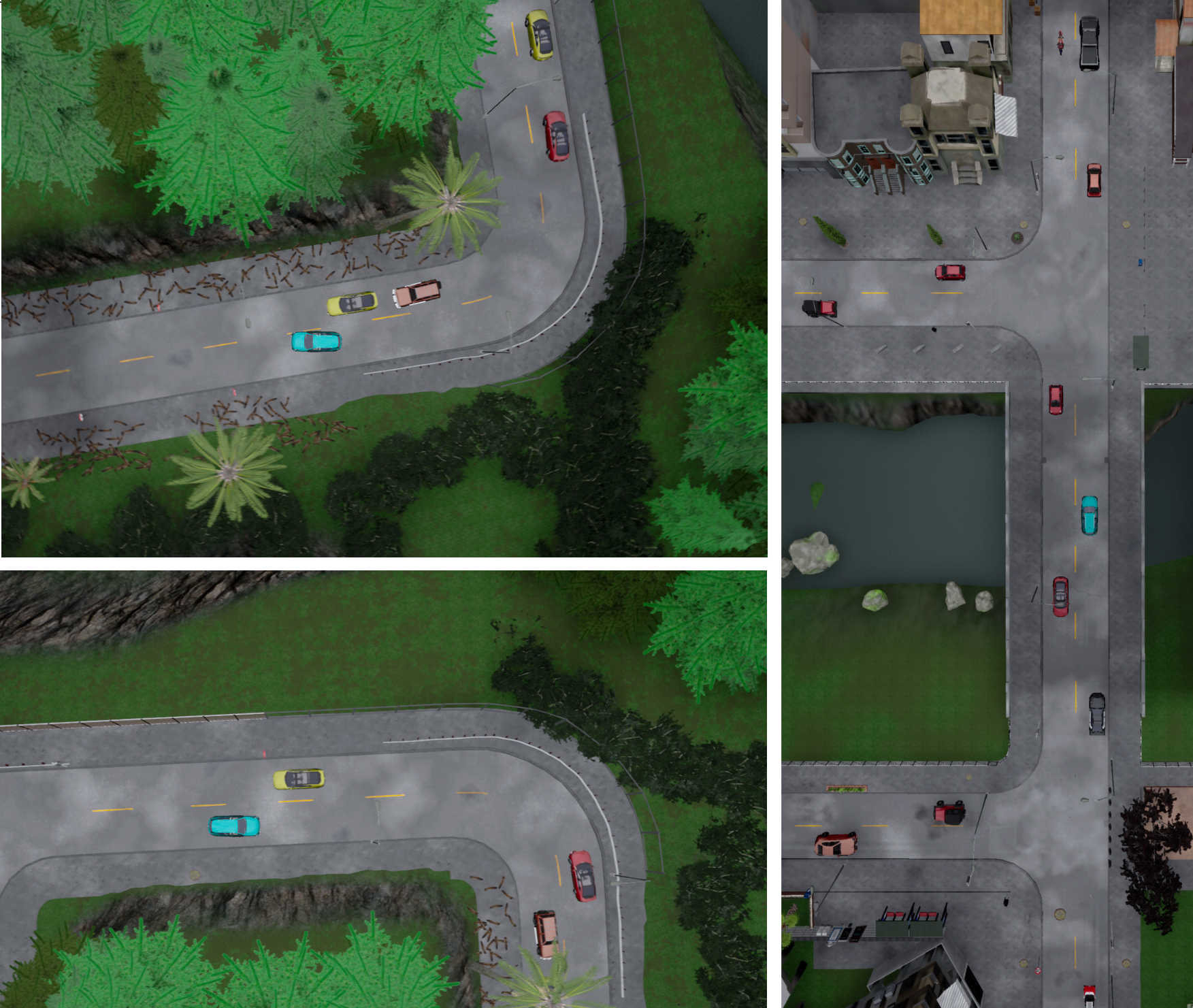}
\caption{Representative example scenarios from the CARLA \textsc{LargeScale} benchmarking dataset, showing a variety of conditions like handling moving vehicles, overtaking static vehicles, road stretches and junctions in \texttt{Town01}.}
\label{fig:dataset-examples}
\end{figure}

We trained and benchmarked \PILOT 
using CARLA simulator (v\,0.9.10)~\cite{carla},
where we can realise complex interactions with, and between, other vehicles that would be hard to generate by synthetically perturbing a scenario. To that end, we obtained 20,604 planning problems from randomly generated scenarios in \texttt{Town02} with up to 40 non-ego vehicles controlled by CARLA's \texttt{Autopilot}. These problems are then solved using \obp to get the base dataset $\mathcal{D}_{0}$. We trained \PILOT using Algorithm~\ref{alg:training}, randomly spawning the \textit{ego} and other vehicles in new simulations. For benchmarking, we generated a dataset of 1,000 problems in \texttt{Town01}, with representative example problems
 shown in Fig.~\ref{fig:dataset-examples}. We refer to this dataset as \textsc{LargeScale} to differentiate it from the one used in Sec.~\ref{sec:q4}.

\subsection{\PILOT vs. \obp}
\label{sec:q1}

We compare the quality of the plans produced by \PILOT and \obp with two metrics:
\begin{itemize}
\item \textit{Solving times} (s) -- the time required to initialise the efficient NLP stage (using the MILP stage in \obp, and using the neural network for \PILOT), NLP solver runtime after initialisation, and the total time. Lower solving time is better. 
\item \textit{Cost} -- the  $\mathcal{J}_{\text{2s-OPT}}$ cost of NLP output upon convergence as in Eq.~\ref{eq:obp-optim} (Appendix~\ref{apx:nlp}), reflecting the quality of the final solution, where lower cost values are better.
\end{itemize}

\bgroup
\def\arraystretch{1.5}
\begin{table}[t]
    \caption{\PILOT vs. \obp: solving time and cost ($\mathcal{J}_{\text{2s-OPT}}$) on 962/1,000 problems where both \PILOT \& \obp converge (mean $\pm$ standard deviation over a varied set of driving problems)}
    \centering
    \begin{tabular}{p{0.90cm}|c|c|c|p{1.3cm}}
    \hline
    \multirow{2}{*}{Planner}	 & 	\multicolumn{3}{c|}{Time~(s)} & 	\multirow{2}{*}{Cost}\\
        \cline{2-4}
        	 & 	Initialisation	 & 	NLP & 	Total 	 & 	\\
        \hline
        \textbf{\PILOT} & $\textbf{0.02}\pm \textbf{0.00}$	 & 	$\textbf{0.10}\pm \textbf{0.15}$ & 	$\textbf{0.12}\pm \textbf{0.15}$	 & 	$0.58\pm 0.69$	\\
        \obp	 & $0.70\pm 1.25$	 & 	$0.17\pm 0.23$	 & 	$0.87\pm 1.31$	 & 	$\textbf{0.57}\pm \textbf{0.68}$	\\
        \hline
    \end{tabular}
    
    \label{table:test-set-results}
\end{table}
\egroup

We report the value of these metrics in the \textsc{LargeScale} benchmarking dataset in Table~\ref{table:test-set-results},  vindicating our approach of combining an imitation learning network with an optimiser to produce an efficient, safe planner. \PILOT shows a clear advantage in runtime efficiency when compared to \obp, leading to savings of $\sim\,$\textbf{86\%} in total runtime, with no significant deterioration in solution quality ($\sim\,$5\% drop). 

\subsection{\PILOT vs. $\networkobp$ }
\label{sec:q2}

\begin{figure}[t]
\centering
\includegraphics[width=0.49\textwidth]{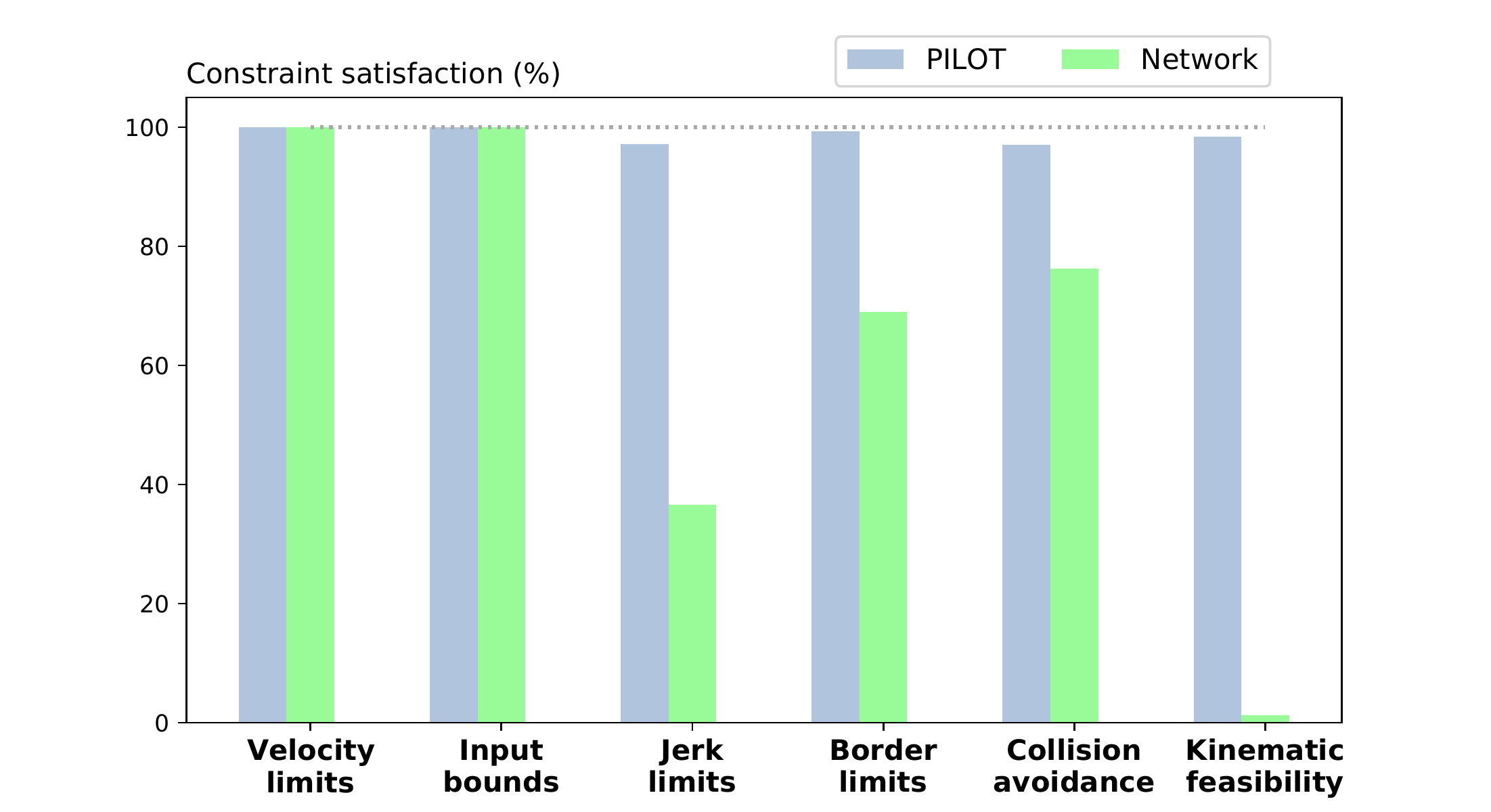}
\caption{\PILOT vs. $\networkobp$: constraint satisfaction percentages in the \textsc{LargeScale} benchmarking dataset of 1,000 problems.}
\label{fig:N_pilot_constraint_satisfaction}
\end{figure}

We showcase the  advantages of having an efficient optimiser rectifying the network mistakes at inference time by comparing the trajectories obtained by \PILOT to those obtained by $\networkobp$, \PILOT's trained imitation network, when used by itself as a planner. We show in Fig.~\ref{fig:N_pilot_constraint_satisfaction} the satisfaction rate of the constraints as defined in Sec.~\ref{sec:obp-details} in the \textsc{LargeScale} benchmark dataset.

As can be observed, $\networkobp$ struggles to reach the constraint satisfaction levels of \PILOT, particularly with the equality   kinematic feasibility constraints. While this particular kind of constraint could be addressed with additional kinematic output layers in the network~\cite{cui20}, \PILOT provides a simpler and a more general approach that improves the satisfiability of most constraints.

\subsection{Efficient optimiser initialisation ablation}
\label{sec:q3}

We present an ablation study on the quality of the imitation network as an initialisation to the efficient NLP optimiser, compared to simple, heuristic alternatives. In particular we consider: \textbf{None} -- an initialisation which sets \textit{ego} position, yaw and speed to zero at all timesteps; \textbf{ConstVel} -- a constant velocity initialisation that maintains the \textit{ego}'s heading; and \textbf{ConstAccel/ConstDecel} -- constant acceleration and deceleration initialisations for which the speed is changed with a constant rate until it reaches the speed limit or 0, respectively.

We compare the alternatives, relative to the original \obp MILP stage initialisation, in the \textsc{LargeScale} benchmarking dataset with three metrics:
\begin{itemize}
    \item \textit{$\Delta$ NLP solving time} and \textit{$\Delta$ NLP cost}-- we report the average difference in solving time (relative to MILP) and the percentage change in the cost of the output trajectory compared to MILP in the problems that both the initialisation method and \obp solved.
    \item \textit{Percentage of solved problems} -- constrained, non-linear optimisation in general is not guaranteed to converge to a feasible solution, hence the quality of an initialisation would be reflected in a higher percentage of solved problems. We report the percentage of solved problems  out of the problems that \obp solved.
\end{itemize}

Results in Table~\ref{table:ablation-results2} show that \PILOT's neural network initialisation produces trajectories that are easier to optimise (reduced NLP solving time) with only a small averaged increase in the final cost  compared to MILP. ConstAccel has a slight advantage in NLP cost on the problems it solves, but solves far fewer and takes significantly longer to converge.

\bgroup
\def\arraystretch{1.5}
\begin{table}[t]
    \caption{$\nu$ initialisation ablation: comparison of each method w.r.t. \obp on mean solving time/NLP cost (in problems solved by both \obp \& initialisation) and convergence percentage.}
    \centering
    \begin{tabular}{l|p{1.7cm}|p{1.5cm}||p{1.2cm}}
        \hline
        Initialisation	 &  $\Delta$~NLP solve time (s) & $\Delta$~NLP cost (\%) & 	Converged (\%)\\
        \hline
        None             & +0.66   &  +9.3\%           & 89.5\% \\
        ConstVel         & +0.18   &  +2.9\%           & 95.3\% \\
        ConstAccel       & +0.44   &  \textbf{-0.1\%}   & 91.7\% \\
        ConstDecel       & +0.35   &  +8.9\%           & 95.3\% \\ 
        $\networkobp$ \textbf{(\PILOT)} & \textbf{-0.07}  & +2.3\%  & \textbf{96.8\%} \\ 
        \hline
        
        MILP (\obp)              & \hspace{0.25cm}-     &   \hspace{0.25cm}- & ${99.2\%}$\\
        \hline
    \end{tabular}
    \label{table:ablation-results2}
\medskip
\end{table}
\egroup

\subsection{\PILOT vs. CPN }
\label{sec:q4}

We showcase the advantages of our framework by comparing it to an optimiser-free alternative: CPN~\cite{cpn}, a state-of-the-art method that trains a neural network directly with a loss function that approximates the optimiser cost function. 

Attempts to train CPN na\"ively on \textsc{LargeScale} failed to result in an effective network, leading to the more elaborate training procedure discussed in Appendix~\ref{apx:cpn}. Thus, to facilitate a fair comparison between \PILOT and CPN, we created a simpler dataset in Carla's \texttt{Town02} (\textsc{SmallScale}), with 20,000 problems that are limited to up to 3 static vehicles on a straight stretch of road. We use a dataset of 1,000 problems randomly generated in the same way for benchmarking.

\begin{figure}[t]
\centering
\includegraphics[width=0.49\textwidth]{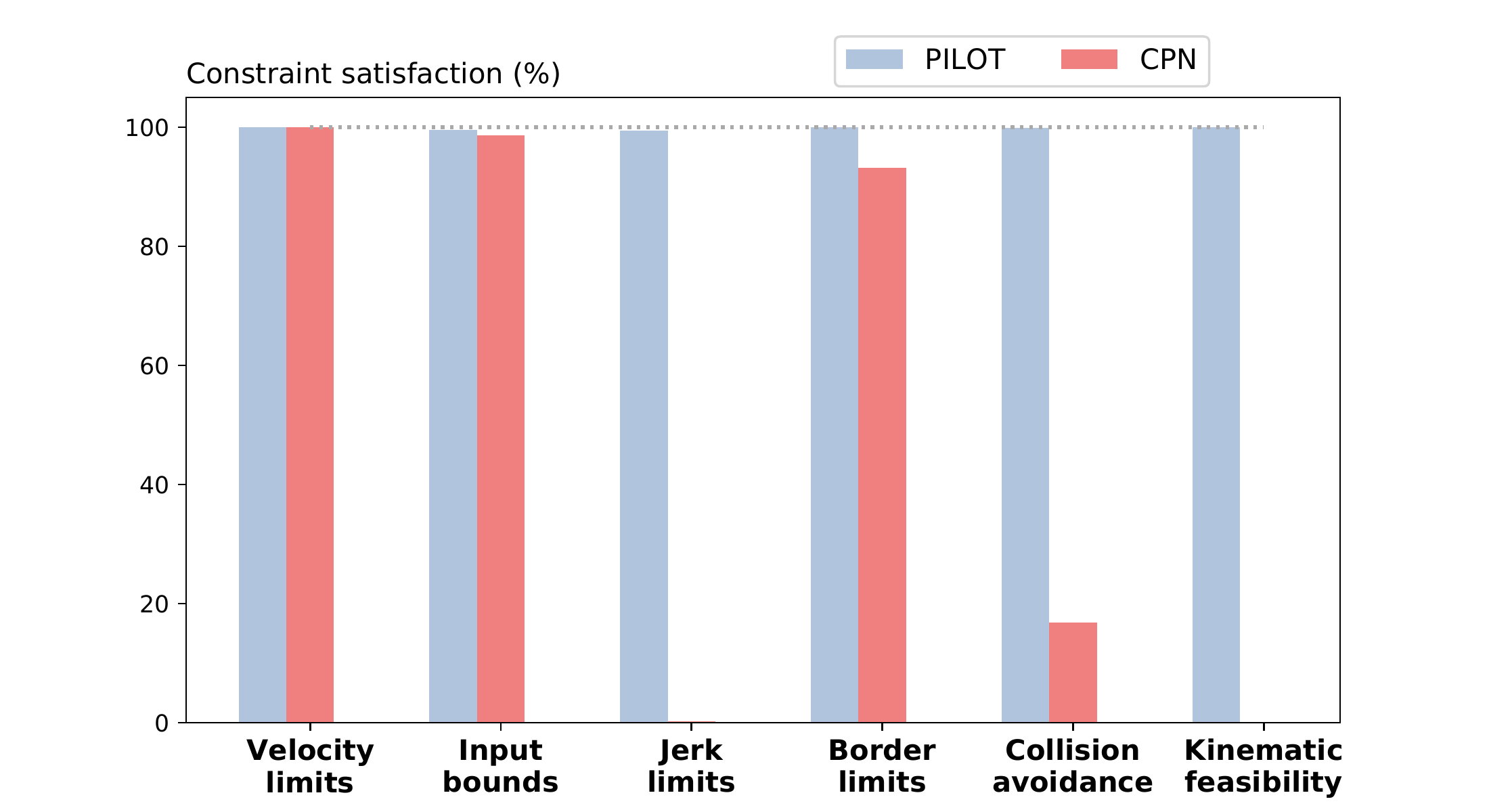}
\caption{\PILOT vs. CPN: constraint satisfaction percentages in \textsc{SmallScale} benchmarking dataset of 1,000 problems.}
\label{fig:cpn_pilot_constraint_satisfaction}
\end{figure}

Fig.~\ref{fig:cpn_pilot_constraint_satisfaction} shows a bar plot of constraint satisfaction rates in \textsc{SmallScale} benchmark dataset. CPN fails to guarantee kinematic feasibility, collision safety and comfort requirements in the output. On the other hand, \PILOT is guaranteed to satisfy these requirements when the efficient optimiser $\nu$ converges --  99.4\% of the problems in this dataset.

\section{Discussion and Conclusion}
\label{sec:discussion}

We now review the questions posed in Sec.~\ref{sec:experiments} in light of the experimental results. We demonstrated in Sec.~\ref{sec:q1} the effectiveness of \PILOT in replacing an \textit{expensive-to-run} optimiser (e.g., \obp) by showing  a reduction of nearly $7\times$ in runtime compared to the optimiser while only suffering a marginal increase in final cost. As mentioned in Sec.~\ref{sec:background}, the expensive-to-run optimiser we use here builds on the fast framework of Schwarting \textit{et al.}~\cite{schwarting2017safe} that  can re-plan at 10Hz but suffers from local convergence issues.  \obp~\cite{obp} offers an improvement in convergence and quality at the cost of runtime efficiency, with our implementation having a re-plan rate of around 1Hz. \PILOT, when applied to \obp, allows for a re-plan rate of 8Hz, approaching the original speed of~\cite{schwarting2017safe} but with improved convergence and lower cost solutions similar to \obp.

Our training procedure is an imitation learning paradigm with online dataset augmentation using the expensive-to-run optimiser as the expert. This could be interpreted as a technique of policy distillation~\cite{policy-distillation}, replacing the sophisticated expert with a much more efficient proxy. The generalisation power of the expert is maintained to some extent through the efficient optimiser stage that actively tries to satisfy the same constraints as the expert. The initialisation ablation of the efficient optimiser presented in Sec.~\ref{sec:q3} showcases the benefits and the quality of  trained imitation network when compared to simple alternatives. The simplicity of our training paradigm is corroborated further by the comparison to CPN in Sec.~\ref{sec:q4}. The nature of the optimisation of CPN training for complex formulations with many constraints as we have in \obp results in a difficult training process, requiring careful fine-tuning (see Appendix~\ref{apx:cpn}). \PILOT, on the other hand, relies on the solutions of the expert with a simple $L_2$ loss, limiting the need for fine-tuning. 

We justify the design of our inference procedure in Sec.~\ref{sec:q2}, showing that \PILOT's efficient optimiser effectively corrects the network output, leading to safer, constraint satisfying solutions. Moreover, the efficient optimiser  operates on the full-length, long-horizon trajectory that is produced by the network, in contrast to existing approaches in which the optimisation at inference time is restricted to a limited horizon~\cite{sun2018fast}. In the case of~\cite{sun2018fast} in particular, the short-term MPC problem is highly conditioned on the network's output, which has the potential of creating sub-optimal, and even unsafe, solutions if the network yields a poor result.

The complexity of the expert optimiser and the cost of running it  within our framework influences only the training phase of the imitation network and has no effect on the inference phase. Thus, in the future we are interested in exploring more advanced experts, e.g., returning the solution with the minimum cost using an ensemble of initialisations~\cite{friesen2016recursive}. Furthermore, one could investigate applying conditional imitation learning~\cite{codevilla2018end} and other loss functions, e.g. $L_1$~\cite{chen2020learning}, to improve further the quality of the initialisation  provided by the network and bridge the existing gap between the expert and efficient optimisers. 

\appendix

\subsection{Nonlinear programming problem formulation}
\label{apx:nlp}

Following the definition from~\cite{obp}, we take $\Delta t$ to be the timestep between states, $N$ to be the desired plan length, and we assume the discretised kinematic bicycle model $\x_{k+1} = f_{\Delta t}(\x_k, \cnt_k)$  where $\x_k = (x_k, y_k, \theta_k, v_k)$ is \textit{ego} state (pose and speed) and $\cnt_k = (a_k, \delta_k)$ is the control inputs (acceleration and steering) applied to the \textit{ego} at step $k$. The goal of the \obp framework is to solve the following constrained optimisation problem:
\begin{equation}
\begin{aligned}
& \underset{\x_{1:N}, \cnt_{0:N-1}}{\text{argmin}}
& & \mathcal{J}_{\text{2s-OPT}}(\x_{1:N}, \cnt_{0:N-1}) \\
& \text{\hspace{2em} s.t.} & & \x_{k+1} = f_{\Delta t}(\x_k, \cnt_k)\\
& & & 0 \leq v_{\min} \leq v_k \leq v_{\max} \\
& & & |\delta_k| \leq \delta_{\max}\\
& & & a_{\min} \leq a_k \leq a_{\max} \\
& & & |a_{k+1} - a_k| \leq \dot{a}_{\max}\\
& & & |\delta_{k+1} - \delta_k| \leq \dot{\delta}_{\max} \\
& & & \mathcal{E}(\x_k) \cap \left(\left[\mathbb{R}^2\setminus\mathcal{B}\right] \cup \mathcal{S}^{1:w}_k\right) = \emptyset\text{, }\forall k\\
\end{aligned}
\label{eq:obp-optim}
\end{equation}
where $v_{\min}$ is the minimum desired speed, $v_{\max}$ is the road's speed limit, $\delta_{\max}$ is maximum allowed steering input, $[a_{\min}, a_{\max}]$ is the allowed range for acceleration/deceleration commands, $\dot{a}_{\max}$ is the maximum allowed jerk, $\dot{\delta}_{\max}$ is the maximum allowed angular jerk. Additionally, $\mathcal{B} \subset \mathbb{R}^2$ is the driveable surface that is safe to drive based on the layout, $\mathcal{S}^{1:w}_{1:N} \subset \mathbb{R}^{2\times N}$ are unions of elliptical areas that encompass the $w$ road users, $\mathcal{S}^{1:w}_{k}$, for timesteps $k \in \{1, ..., N\}$, $\mathcal{E}(\x_k) \subset \mathbb{R}^2$ is the area the \textit{ego} occupies at step $k$ with, and $\mathcal{J}_{\text{2s-OPT}}$ is a cost function comprising a linear combination of quadratic terms of comfort (reduced acceleration and jerk) and progress (longitudinal and lateral tracking of the reference path, as well as speed)~\cite{obp}. In \obp, the \textit{ego}'s area $\mathcal{E}(\x_k)$ is approximated   by its corners, so that the intersection with the driveable surface -- delimited by its borders which are defined as $C^2$ functions -- and road user ellipses can be computed in closed form~\cite{obp}.

The cost function to optimise is defined as 
\begin{equation}
\label{eq:cost}
\mathcal{J}_{\text{2s-OPT}}(\x_{1:N}, \cnt_{0:N-1}) = \sum_{k=0}^{N}
\sum_{\iota \in \mathcal{I}} \omega_\iota \theta_\iota(\x_k, \cnt_k)
\end{equation}
where $\omega_\iota\in\mathbb{R}$ are scalar weights, and $\theta_\iota(\mathbf{z}_k, \mathbf{u}_k)$ are soft  constraints that measure deviation from the desired speed ($\omega_v$), the reference path ($\omega_y$) and the end target location ($\omega_x$), and that control the norms of acceleration and steering control inputs ($\omega_a$ and $\omega_\delta$). We fine-tune the parameters of the optimisation using grid-search in the parameter space.

The parameters of the optimisation are in Table~\ref{tab:params}.

\begin{table}[t]
\caption{NLP weights and parameters}

\centering
\bgroup
\def\arraystretch{1.3}
\resizebox{0.5\textwidth}{!}{%
\begin{tabular}{l|l||l|l||l|l}
Parameter & Value & Parameter & Value & Parameter & Value \\ \hline
$L$              & $4.8$ $m$      & $\dot{\delta}_{\max}$ & $0.18$ $rad/s^2$ & $\omega_x$ & $0.1$  \\ 
$\delta_{\max}$  & $0.45$ $rad/s$ & $v_{\max}$            & $10$ $m/s$       & $\omega_v$ & $2.5$  \\
$a_{\min}$       & $-3$ $m/s^2$   & $v_{\min}$            & $0$ $m/s$        & $\omega_y$ & $0.05$ \\
$a_{\max}$       & $3$ $m/s^2$    & $\omega_{\delta}$     & $2.0$            & $\omega_a$ & $1.0$  \\
$\dot{a}_{\max}$ & $0.5$ $m/s^3$  &                       &                  &            &        \\
\end{tabular}
}
\egroup
\label{tab:params}
\end{table}

\subsection{\PILOT for \obp: deep neural network architecture}
\label{sec:app_network_architecture}

\begin{figure}[ht!]
\centering
\includegraphics[width=0.5\textwidth]{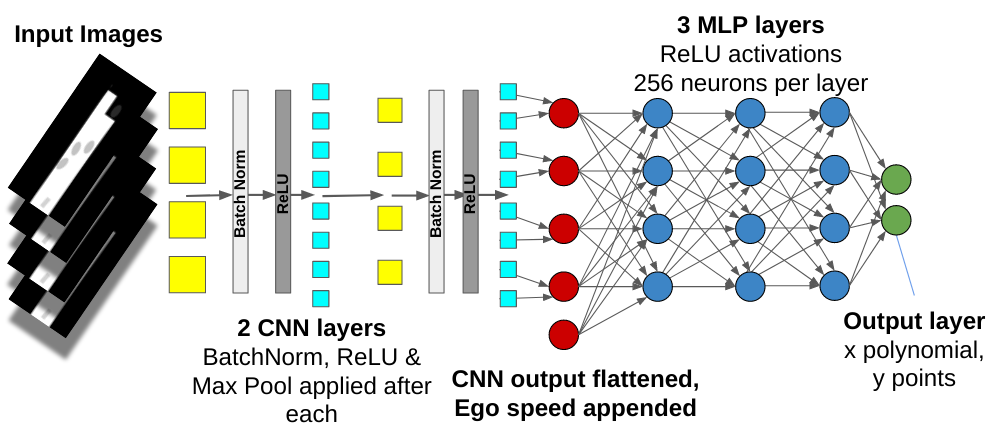}
\caption{\PILOT for \obp network architecture.}
\label{fig:pilot_NN}
\end{figure}

\subsection{Output transformation checks} \label{apx:checks}
The network produces a sequence of spatial positions, then the rest of the required input of the efficient optimiser are computed from that sequence. A number of checks of upper and lower limits are applied to tame abnormalities in the network output and to improve the input to the optimiser:
\begin{itemize}
    \item Velocity limits: $v_k \in[0,v_{\max}]$
    \item Acceleration/deceleration limits: $a_k\in[a_{\min},  a_{\max}]$
    \item Maximum jerk limit: $|a_{k+1} - a_k| \leq \dot{a}_{\max}$
    \item Maximum steering angle limit: $|\delta_k| \leq \delta_{\max}$
    \item Maximum angular jerk limit: $|\delta_{k+1} - \delta_k| \leq \dot{\delta}_{\max}$
\end{itemize}

\subsection{CPN baseline training procedure}
\label{apx:cpn}
After many failed attempts at training with all constraints from a random initialisation, we applied a curriculum learning approach~\cite{curriculumlearning}, sequentially introducing constraints and tuning their weights with each introduction. This approach does not  scale well to complex constraint sets as it requires expert knowledge of the constraints.

In our case, all terms of $\mathcal{J}_{\text{2s-OPT}}$ satisfy differentiability. To make the hard constraint terms differentiable, we approximate them with ReLUs that penalise constraint violation as in~\cite{cpn}. The ReLUs have large gradients to ensure they are prioritised over soft constraints. As the hard constraints have different units, they require normalising to ensure the cost function and gradient used for training reflect this.

\bibliographystyle{IEEEtran}
\bibliography{refs}

\begin{thebibliography}{10}
\providecommand{\url}[1]{#1}
\csname url@samestyle\endcsname
\providecommand{\newblock}{\relax}
\providecommand{\bibinfo}[2]{#2}
\providecommand{\BIBentrySTDinterwordspacing}{\spaceskip=0pt\relax}
\providecommand{\BIBentryALTinterwordstretchfactor}{4}
\providecommand{\BIBentryALTinterwordspacing}{\spaceskip=\fontdimen2\font plus
\BIBentryALTinterwordstretchfactor\fontdimen3\font minus
  \fontdimen4\font\relax}
\providecommand{\BIBforeignlanguage}[2]{{%
\expandafter\ifx\csname l@#1\endcsname\relax
\typeout{** WARNING: IEEEtran.bst: No hyphenation pattern has been}%
\typeout{** loaded for the language `#1'. Using the pattern for}%
\typeout{** the default language instead.}%
\else
\language=\csname l@#1\endcsname
\fi
#2}}
\providecommand{\BIBdecl}{\relax}
\BIBdecl

\bibitem{batkovic2019real}
I.~Batkovic, M.~Zanon, M.~Ali, and P.~Falcone, ``Real-time constrained
  trajectory planning and vehicle control for proactive autonomous driving with
  road users,'' in \emph{Proceedings of the European Control Conference}.\hskip
  1em plus 0.5em minus 0.4em\relax IEEE, 2019, pp. 256--262.

\bibitem{chen2019autonomous}
J.~Chen, W.~Zhan, and M.~Tomizuka, ``Autonomous driving motion planning with
  constrained iterative {LQR},'' \emph{IEEE Transactions on Intelligent
  Vehicles}, vol.~4, no.~2, pp. 244--254, 2019.

\bibitem{schwarting2017safe}
W.~Schwarting, J.~Alonso-Mora, L.~Paull, S.~Karaman, and D.~Rus, ``Safe
  nonlinear trajectory generation for parallel autonomy with a dynamic vehicle
  model,'' \emph{IEEE Transactions on Intelligent Transportation Systems},
  vol.~19, no.~99, 2017.

\bibitem{obp}
F.~Eiras, M.~Hawasly, S.~V.~Albrecht, and S.~Ramamoorthy, ``A two-stage
  optimization-based motion planner for safe urban driving,'' \emph{IEEE
  Transactions on Robotics}, pp. 1--13, 2021, to appear.

\bibitem{sadat2020perceive}
A.~Sadat, S.~Casas, M.~Ren, X.~Wu, P.~Dhawan, and R.~Urtasun, ``Perceive,
  predict, and plan: Safe motion planning through interpretable semantic
  representations,'' in \emph{European Conference on Computer Vision}.\hskip
  1em plus 0.5em minus 0.4em\relax Springer, 2020, pp. 414--430.

\bibitem{bojarski2016end}
M.~Bojarski, D.~Del~Testa, D.~Dworakowski, B.~Firner, B.~Flepp, P.~Goyal, L.~D.
  Jackel, M.~Monfort, U.~Muller, J.~Zhang \emph{et~al.}, ``End to end learning
  for self-driving cars,'' \emph{arXiv:1604.07316}, 2016.

\bibitem{pan2020imitation}
Y.~Pan, C.-A. Cheng, K.~Saigol, K.~Lee, X.~Yan, E.~A. Theodorou, and B.~Boots,
  ``Imitation learning for agile autonomous driving,'' \emph{International
  Journal of Robotics Research}, vol.~39, no. 2-3, pp. 286--302, 2020.

\bibitem{hawke2020urban}
J.~Hawke, R.~Shen, C.~Gurau, S.~Sharma, D.~Reda, N.~Nikolov, P.~Mazur,
  S.~Micklethwaite, N.~Griffiths, A.~Shah \emph{et~al.}, ``Urban driving with
  conditional imitation learning,'' in \emph{Proceedings of the IEEE
  International Conference on Robotics and Automation (ICRA)}.\hskip 1em plus
  0.5em minus 0.4em\relax IEEE, 2020, pp. 251--257.

\bibitem{chen2020learning}
D.~Chen, B.~Zhou, V.~Koltun, and P.~Kr{\"a}henb{\"u}hl, ``Learning by
  cheating,'' in \emph{Proceedings of the Conference on Robot Learning}.\hskip
  1em plus 0.5em minus 0.4em\relax PMLR, 2020, pp. 66--75.

\bibitem{mirman2018differentiable}
M.~Mirman, T.~Gehr, and M.~Vechev, ``Differentiable abstract interpretation for
  provably robust neural networks,'' in \emph{Proceedings of the International
  Conference on Machine Learning}, 2018, pp. 3578--3586.

\bibitem{parot}
E.~W. Ayers, F.~Eiras, M.~Hawasly, and I.~Whiteside, ``{P}a{R}o{T}: A practical
  framework for robust deep neural network training,'' in \emph{NASA Formal
  Methods}.\hskip 1em plus 0.5em minus 0.4em\relax Springer, 2020, pp. 63--84.

\bibitem{liu2019algorithms}
C.~Liu, T.~Arnon, C.~Lazarus, C.~Barrett, and M.~J. Kochenderfer, ``Algorithms
  for verifying deep neural networks,'' \emph{arXiv preprint arXiv:1903.06758},
  2019.

\bibitem{hanna21}
J.~P. Hanna, A.~Rahman, E.~Fosong, F.~Eiras, M.~Dobre, J.~Redford,
  S.~Ramamoorthy, and S.~V. Albrecht, ``Interpretable goal recognition in the
  presence of occluded factors for autonomous vehicles,'' in \emph{Proceedings
  of the IEEE/RSJ International Conference on Intelligent Robots and Systems
  (IROS)}.\hskip 1em plus 0.5em minus 0.4em\relax IEEE, 2021.

\bibitem{albrecht2020integrating}
S.~V. Albrecht, C.~Brewitt, J.~Wilhelm, B.~Gyevnar, F.~Eiras, M.~Dobre, and
  S.~Ramamoorthy, ``Interpretable goal-based prediction and planning for
  autonomous driving,'' in \emph{Proceedings of the IEEE International
  Conference on Robotics and Automation (ICRA)}.\hskip 1em plus 0.5em minus
  0.4em\relax IEEE, 2021.

\bibitem{DeCastro2020InterpretablePF}
J.~A. DeCastro, K.~Leung, N.~Ar{\'e}chiga, and M.~Pavone, ``Interpretable
  policies from formally-specified temporal properties,'' in \emph{Proceedings
  of the International Conference on Intelligent Transportation Systems
  (ITSC)}, 2020.

\bibitem{paden2016survey}
B.~Paden, M.~{\v{C}}{\'a}p, S.~Z. Yong, D.~Yershov, and E.~Frazzoli, ``A survey
  of motion planning and control techniques for self-driving urban vehicles,''
  \emph{IEEE Transactions on intelligent vehicles}, vol.~1, no.~1, pp. 33--55,
  2016.

\bibitem{policy-distillation}
A.~A. Rusu, S.~G. Colmenarejo, {\c{C}}.~G{\"{u}}l{\c{c}}ehre, G.~Desjardins,
  J.~Kirkpatrick, R.~Pascanu, V.~Mnih, K.~Kavukcuoglu, and R.~Hadsell, ``Policy
  distillation,'' in \emph{Proceedings of the International Conference on
  Learning Representations ({ICLR})}, 2016.

\bibitem{dagger}
S.~Ross, G.~Gordon, and D.~Bagnell, ``A reduction of imitation learning and
  structured prediction to no-regret online learning,'' in \emph{Proceedings of
  the International Conference on Artificial Intelligence and Statistics},
  2011, pp. 627--635.

\bibitem{carla}
A.~Dosovitskiy, G.~Ros, F.~Codevilla, A.~Lopez, and V.~Koltun, ``{CARLA}: {An}
  open urban driving simulator,'' in \emph{Conference on Robot Learning}, ser.
  Proceedings of Machine Learning Research, vol.~78.\hskip 1em plus 0.5em minus
  0.4em\relax PMLR, 2017, pp. 1--16.

\bibitem{pomerleau1989alvinn}
D.~A. Pomerleau, ``Alvinn: An autonomous land vehicle in a neural network,'' in
  \emph{Advances in neural information processing systems}, 1989, pp. 305--313.

\bibitem{codevilla2019exploring}
F.~Codevilla, E.~Santana, A.~M. L{\'o}pez, and A.~Gaidon, ``Exploring the
  limitations of behavior cloning for autonomous driving,'' in
  \emph{Proceedings of the IEEE International Conference on Computer Vision},
  2019, pp. 9329--9338.

\bibitem{filos2020can}
A.~Filos, P.~Tigas, R.~McAllister, N.~Rhinehart, S.~Levine, and Y.~Gal, ``Can
  autonomous vehicles identify, recover from, and adapt to distribution
  shifts?'' in \emph{Proceedings of the International Conference on Machine
  Learning (ICML)}, 2020.

\bibitem{chen2019deep}
J.~Chen, B.~Yuan, and M.~Tomizuka, ``Deep imitation learning for autonomous
  driving in generic urban scenarios with enhanced safety,'' in \emph{IEEE/RSJ
  International Conference on Intelligent Robots and Systems (IROS)}, 2019, pp.
  2884--2890.

\bibitem{bansal2018chauffeurnet}
M.~Bansal, A.~Krizhevsky, and A.~Ogale, ``Chauffeurnet: Learning to drive by
  imitating the best and synthesizing the worst,'' in \emph{Robotics: Science
  and Systems}, 2019.

\bibitem{lee2018safe}
K.~Lee, K.~Saigol, and E.~A. Theodorou, ``Safe end-to-end imitation learning
  for model predictive control,'' \emph{arXiv preprint arXiv:1803.10231}, 2018.

\bibitem{sun2018fast}
L.~Sun, C.~Peng, W.~Zhan, and M.~Tomizuka, ``A fast integrated planning and
  control framework for autonomous driving via imitation learning,'' in
  \emph{Proceedings of the Dynamic Systems and Control Conference},
  vol.~3.\hskip 1em plus 0.5em minus 0.4em\relax ASME, 2018.

\bibitem{acerbo2020safe}
F.~S. Acerbo, H.~Van~der Auweraer, and T.~D. Son, ``Safe and computational
  efficient imitation learning for autonomous vehicle driving,'' in
  \emph{Proceedings of the American Control Conference (ACC)}.\hskip 1em plus
  0.5em minus 0.4em\relax IEEE, 2020, pp. 647--652.

\bibitem{cpn}
W.~Zhan, J.~Li, Y.~Hu, and M.~Tomizuka, ``Safe and feasible motion generation
  for autonomous driving via constrained policy net,'' in \emph{Conference of
  the IEEE Industrial Electronics Society}, 2017, pp. 4588--4593.

\bibitem{floudas2013state}
C.~A. Floudas and P.~M. Pardalos, \emph{State of the art in global
  optimization: computational methods and applications}.\hskip 1em plus 0.5em
  minus 0.4em\relax Springer Science \& Business Media, 2013, vol.~7.

\bibitem{nocedal2006numerical}
J.~Nocedal and S.~Wright, \emph{Numerical optimization}.\hskip 1em plus 0.5em
  minus 0.4em\relax Springer Science \& Business Media, 2006.

\bibitem{ipopt}
A.~W{\"a}chter and L.~T. Biegler, ``On the implementation of an interior-point
  filter line-search algorithm for large-scale nonlinear programming,''
  \emph{Mathematical programming}, vol. 106, no.~1, pp. 25--57, 2006.

\bibitem{forcesnlp}
A.~Zanelli, A.~Domahidi, J.~Jerez, and M.~Morari, ``Forces nlp: an efficient
  implementation of interior-point methods for multistage nonlinear nonconvex
  programs,'' \emph{International Journal of Control}, pp. 1--17, 2017.

\bibitem{friesen2016recursive}
A.~L. Friesen and P.~Domingos, ``Recursive decomposition for nonconvex
  optimization,'' in \emph{Proceedings of the International Joint Conference on
  Artificial Intelligence}.\hskip 1em plus 0.5em minus 0.4em\relax AAAI, 2015,
  pp. 253--259.

\bibitem{lembono2020memory}
T.~S. Lembono, A.~Paolillo, E.~Pignat, and S.~Calinon, ``Memory of motion for
  warm-starting trajectory optimization,'' \emph{IEEE Robotics and Automation
  Letters}, vol.~5, no.~2, pp. 2594--2601, 2020.

\bibitem{cui20}
H.~Cui, T.~Nguyen, F.-C. Chou, T.-H. Lin, J.~Schneider, D.~Bradley, and
  N.~Djuric, ``Deep kinematic models for kinematically feasible vehicle
  trajectory predictions,'' in \emph{Proceedings of the IEEE International
  Conference on Robotics and Automation (ICRA)}.\hskip 1em plus 0.5em minus
  0.4em\relax IEEE, 2020, pp. 10\,563--10\,569.

\bibitem{codevilla2018end}
F.~Codevilla, M.~M{\"u}ller, A.~L{\'o}pez, V.~Koltun, and A.~Dosovitskiy,
  ``End-to-end driving via conditional imitation learning,'' in
  \emph{Proceedings of the IEEE International Conference on Robotics and
  Automation (ICRA)}.\hskip 1em plus 0.5em minus 0.4em\relax IEEE, 2018, pp.
  4693--4700.

\bibitem{curriculumlearning}
Y.~Bengio, J.~Louradour, R.~Collobert, and J.~Weston, ``Curriculum learning,''
  in \emph{Proceedings of the International Conference on Machine Learning},
  2009, pp. 41--48.

\end{thebibliography}

\end{document}